\documentclass[letterpaper, 10 pt, conference]{ieeeconf}  
\IEEEoverridecommandlockouts                             
\title{\LARGE \bf
Human-Robot Collaborative Carrying of Objects with \\ Unknown Deformation Characteristics 
}

\author{Doganay Sirintuna, Alberto Giammarino, and Arash Ajoudani
\thanks{Human-Robot Interfaces and physical Interaction (HRI$^{2}$) Lab, Istituto Italiano di Tecnologia, Genoa, Italy.}
\thanks{This work was supported in part by the ERC-StG Ergo-Lean (Grant Agreement No.850932), in part by the European Union’s Horizon 2020 research and innovation programme under Grant Agreement No. 871237 (SOPHIA) and No. 101016007 (CONCERT).}
\thanks{The authors thank to Dr. Marta Lorenzini and Mattia Leonori for the fruitful discussions.}}

\usepackage{cite}
\usepackage{amsmath}
\usepackage{amssymb}
\usepackage[pdftex]{graphicx}
\usepackage[dvipsnames]{xcolor}
\usepackage[hyphens]{url}
\usepackage{caption}
\usepackage{algorithm}
\usepackage{algpseudocode}
\usepackage[english]{datetime2}

\DeclareCaptionFont{mysize}{\fontsize{8}{9.6}\selectfont}
\begin{document}

\maketitle
\thispagestyle{empty}
\pagestyle{empty}

\begin{abstract}

In this work, we introduce an adaptive control framework for human-robot collaborative transportation of objects with unknown deformation behaviour. The proposed framework takes as input the haptic information transmitted through the object, and the kinematic information of the human body obtained from a motion capture system to create reactive whole-body motions on a mobile collaborative robot. In order to validate our framework experimentally, we compared its performance with an admittance controller during a co-transportation task of a partially deformable object. We additionally demonstrate the potential of the framework while co-transporting rigid (aluminum rod) and highly deformable (rope) objects. A mobile manipulator which consists of an Omni-directional mobile base, a collaborative robotic arm, and a robotic hand is used as the robotic partner in the experiments. Quantitative and qualitative results of a 12-subjects experiment show that the proposed framework can effectively deal with objects of unknown deformability and provides intuitive assistance to human partners.

\end{abstract}

\section{INTRODUCTION}

Industrial automation is a crucial enabler for improving productivity and quality, while reducing errors and waste, and for preparing a response to the rapidly aging workforce~\cite{krueger2017have}. In this direction, robotic technologies have been increasingly adopted to replace the human workforce in several repetitive and simple operations. Nevertheless, robots of today still lack operational flexibility and intelligence, typical of humans, which limits their applicability in industrial environments where high flexibility is also required (e.g., logistics). Consequently, workers are still involved in physically demanding tasks, that might jeopardize their health and productivity~\cite{dework}. Building robotic technologies that can effectively collaborate with humans represents one viable way to tackle these issues~\cite{kim2019adaptable}.

\begin{figure}
    \centering
    \resizebox{0.7\columnwidth}{!}{\rotatebox{0}{\includegraphics[trim=11cm 2cm 9cm 1.5cm, clip=true]{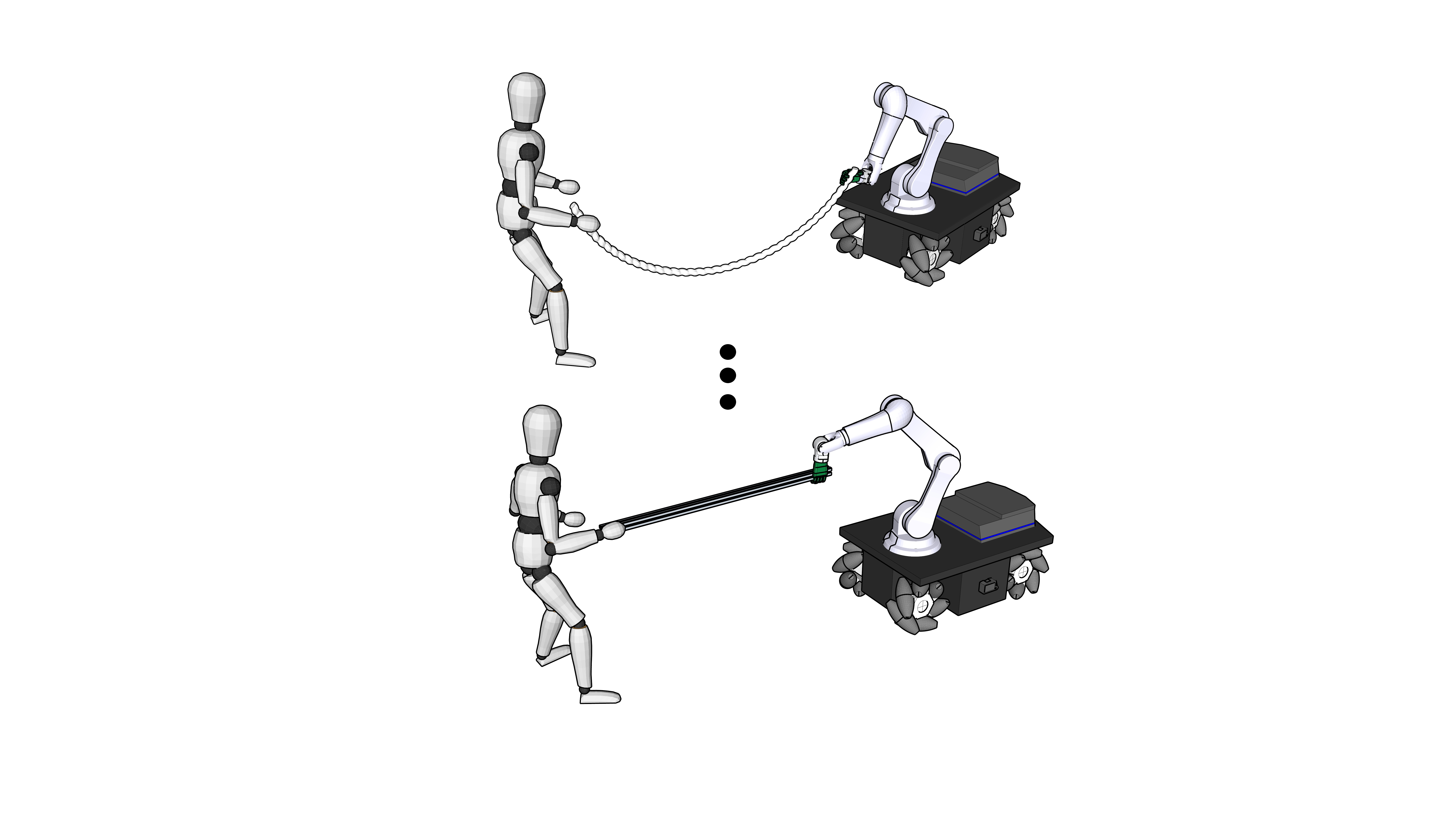}}}
    \caption{We propose an adaptive control framework that enables collaborative carrying of objects with unknown deformation characteristics ranging from highly deformable (e.g., a rope) to rigid (e.g., an aluminum profile).}
    \vspace{-0.7cm}
    \label{fig:digest_figure}
\end{figure}

One frequently encountered task in industrial settings such as factories, warehouses, and construction sites is the transportation of objects. In some cases, this task can be physically demanding while requiring a certain level of adaptability to the surroundings. Moreover, it often requires the cooperation of multiple partners. Instead of an automated solution to this problem, collaborative robots can be used to work together with humans to exploit their cognitive skills in unstructured environments~\cite{Arash,gandarias2022enhancing}.

Three main challenges can be identified in the literature on human-robot co-transportation of objects. First, the robot should effectively share the load, without hindering human intention~\cite{mortl2012role}. Since humans can easily lead the motion during collaboration, the robot should be capable of effectively following them. The second one is the well-known translation/rotation ambiguity~\cite{dumora2012experimental}. Especially, when long objects are carried, humans can hardly control both rotations and translations independently while grasping only one side of the object. This is due to the coupling between different degrees of freedom (DoFs) in the object dynamics~\cite{Takubo2002}. Thirdly, co-transportation of deformable objects represents a major challenge in this research field. Due to the incomplete wrenches transmitted through the object, most haptic-based conventional techniques fall short in this condition.

To address the load sharing problem, in~\cite{ikeura1995variable}, a variable impedance is used to control the robot's end-effector motion, where the damping switches between two discrete values based on end-effector velocity thresholds. Duchaine and Gosselin also adjust the virtual damping of the controller according to the human acceleration/deceleration intention which is predicted by using the time-derivative of the force at the robot's end-effector~\cite{Duchaine,duchaine2009safe}. However, in~\cite{lecours2012variable}, a new criterion for inferring the intended human motion is presented because of the issues related with time-derivative of the force.

Although the works discussed thus far deal with the load sharing problem, they have assumed a follower behaviour of the robot, which limits the level of proactivity it can achieve. Conversely, Bussy et al. estimate the human motion primitives based on the velocity of the object to let the robot actively follow pre-defined sub-motions~\cite{bussy2012proactive}. An approach based on programming by demonstration (PbD) is used in~\cite{evrard2009teaching} to show follower and leader behaviours to a robot during a cooperative lifting task. The two behaviours are encoded through Gaussian Mixture Models (GMMs), and Gaussian Model Regression (GMR) is used to reproduce them during a user study. 

Regarding the translation/rotation ambiguity, there are two main approaches used to tackle this issue~\cite{Karayiannidis}: virtual constraints methods and techniques that consider different motion modes and switch between them. In~\cite{Karayiannidis}, the latter approach is used. The authors define a translation and a rotation mode, and they present two possible methods to switch between them. Instead, a virtual non-holonomic constraint is proposed by Takubo et al. in~\cite{Takubo2002}, consisting of a virtual wheel located at the robot's end-effector. By using this model, lateral movements are not possible and the object is moved as if it was positioned on a cart with fixed passive wheels.

All the previously mentioned works consider only rigid objects. In the literature, a few studies address the challenge of co-manipulating deformable objects~\cite{sanchez2018robotic}. DelPreto and Rus designed a control framework based on EMG information of the upper arm to assist the human operator in a collaborative lifting task~\cite{delpreto2019sharing}. Finally, a controller based on merging force and visual feedback is proposed for the co-transportation of a cloth in~\cite{kruse2015collaborative}. Based on the detected deformed regions of the cloth, this hybrid controller enables the generation of motions that helps the robot to recover tautness of the cloth.

In this article, we present a framework for human-robot co-transportation aiming at tackling two of the three challenges previously mentioned. In particular, the contributions of this paper are as follows:

\begin{itemize}
    \item The design of an adaptive framework that merges haptic and human movement information to generate motion references for a mobile manipulator, enabling it to co-transport objects with unknown deformability while sharing the load effectively during collaboration with a human (Fig.~\ref{fig:digest_figure}).
    \item A multi-subject cross-gender user study to validate the developed framework through quantitative and qualitative results.
\end{itemize}

\section{SYSTEM OVERVIEW}

\begin{figure*}
 \centering
    {\includegraphics[width=0.8\textwidth, trim=0cm 4.8cm 0cm 0cm, clip=true]{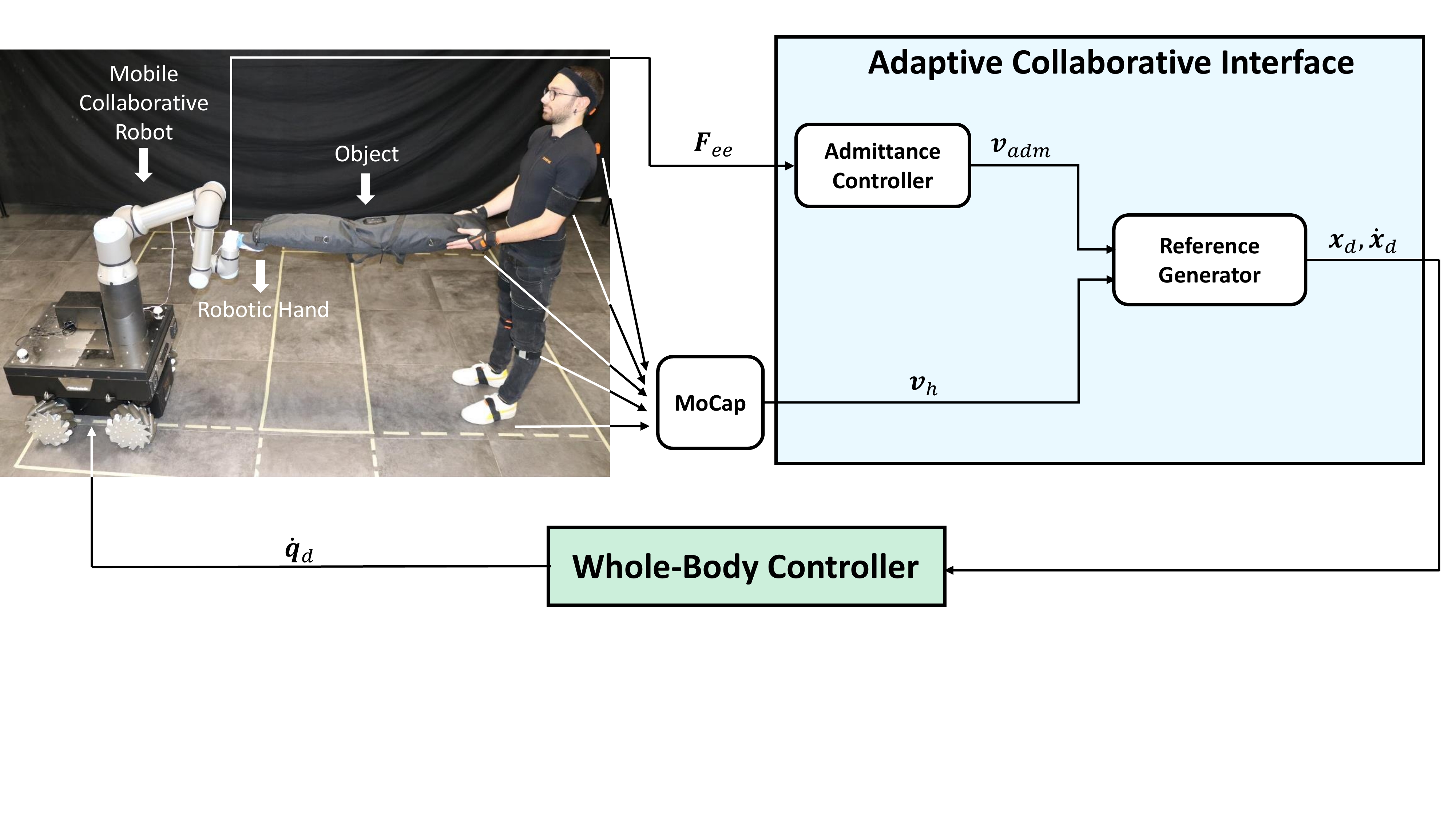}}
    \caption{Our experimental setup and the overall control architecture of the proposed framework.}
    \vspace{-0.7cm}
    \label{fig:system_overview_fig}
\end{figure*}

The hardware and software components of our system are depicted in Fig.~\ref{fig:system_overview_fig}. In this work, we use a mobile base robotic platform, which allows movements over a large workspace. It consists of a $n_b$ DoFs Omni-directional mobile base, a $n_a$ DoFs robotic arm with a Force/Torque (F/T) sensor at its end-effector to measure the wrenches applied at that point of interaction, and an adaptive anthropomorphic robotic hand. In addition, a motion capture (MoCap) system is employed to track movements of the human operator during the co-carrying task. The sensory systems' choice is mainly based on the acquisition of adequate information to enable co-carrying of objects with different deformability (from highly deformable to purely rigid) or those with different perceived deformability in different directions of co-carrying (e.g., a rope that can be relatively rigid when pulled in the constrained direction, generating pure wrenches, but very loose in the opposite direction, generating pure twists).

The overall control architecture of the presented system is composed of two main modules: (1) a Whole-Body Controller which generates desired joint velocities for the mobile base robotic platform and (2) an Adaptive Collaborative Interface that calculates reference inputs for the whole-body controller based on the perceived sensory information. These two components and the interaction between them are explained in detail in the following section.

\section{METHODOLOGY}
\label{sec:methodology}

\subsection{Whole-Body Controller}

The Whole-Body Controller employed on the robot is based on the solution of a Hierarchical Quadratic Program (HQP) composed of two tasks. The formulation of the problem as a HQP allows to exploit the redundancy of the robot, since it has $m=n_b+n_a>6$ DoFs. The cost function of the higher priority task is written as \cite{seraji1990improved} (dependencies are dropped):
\begin{align}
    \mathcal{L}_1 = || \dot{\boldsymbol{x}_d} + \boldsymbol{K}({\boldsymbol{x}_d} -{\boldsymbol{x}}) - \boldsymbol{J}\dot{\boldsymbol{q}}||^2_{\boldsymbol{W}_{1}} + ||k\dot{\boldsymbol{q}}||^2_{\boldsymbol{W}_{2}},
\end{align}
where the vector of the whole-body joint velocities $\boldsymbol{\dot{q}}$ $\in$ $\mathbb{R}^{m}$ is the optimization variable, ${\boldsymbol{J}}$ $\in$ $\mathbb{R}^{6\times m}$ is the whole-body Jacobian, ${\boldsymbol{x}}$ $\in$ $\mathbb{R}^{6}$ is the current end-effector pose, ${\boldsymbol{W}_1}$ $\in$ $\mathbb{R}^{6\times6}$, ${\boldsymbol{W}_2}$ $\in$ $\mathbb{R}^{m\times m}$, and ${\boldsymbol{K}}$ $\in$ $\mathbb{R}^{6\times6}$ are diagonal positive definite matrices and $k$ $\in$ $\mathbb{R}_{>0}$ is the so-called damping factor \cite{chiaverini1992weighted}, which depends on the manipulability index of the arm \cite{deo1995overview, hollerbach1987redundancy, wampler1986manipulator}. Then, the joint velocities of the secondary task are computed as the negative gradient w.r.t. $\boldsymbol{q}$ of (\ref{eq:secondary_task})~\cite{nakanishi2005comparative, wu2021unified}, where ${\boldsymbol{W}_3}$ $\in$ $\mathbb{R}^{m\times m}$ is a diagonal positive semidefinite matrix and ${\boldsymbol{q}_{def}}$ is a default joint configuration: 
\begin{equation}
\label{eq:secondary_task}
    ||\boldsymbol{q}_{def} - \boldsymbol{q}||^2_{\boldsymbol{W}_{3}}.
\end{equation}
These velocities are later projected in the null-space of the primary task.

This formulation allows to obtain the desired whole-body joint velocities $\boldsymbol{\dot{q}}_d$ $\in$ $\mathbb{R}^{m}$ that distribute the movements between base and arm. The primary task guarantees the tracking of the desired end-effector motion ($\boldsymbol{\dot{x}}_d$ and ${\boldsymbol{x}_d}$). On the other hand, the secondary task keeps the arm close to the joints configuration $\boldsymbol{q}_{def}$, which can be chosen in order to ensure a task-related robot posture during the co-carry.

\vspace{-0.1mm}

\subsection{Adaptive Collaborative Interface}

The detailed architecture of our Adaptive Collaborative Interface (ACI) that handles objects with different deformation properties is illustrated in Fig.~\ref{fig:system_overview_fig}. The presented interface can be subdivided into two operational units, namely an \emph{Admittance Controller} which computes a reference velocity based on the force transferred through the object, and a \emph{Reference Generator} that sends the actual reference pose and twist to the Whole-Body Controller based on the combination of the velocity calculated by the Admittance Controller together with the reference velocity produced by the MoCap system.

\subsubsection{Admittance Controller}

In this work, a standard admittance controller is utilized. It implements the following transfer function expressed in Laplace domain:
\begin{equation}
  \boldsymbol{{V}}_{adm}(s) = \frac{\boldsymbol{F}_H(s)}{\boldsymbol{M}_{adm}s+\boldsymbol{D}_{adm}},
\end{equation}

\noindent where $\boldsymbol{M}_{adm}$ and $\boldsymbol{D}_{adm}$  $\in$ $\mathbb{R}^{3\times3}$ are the desired mass and damping matrices, $s$ is the Laplace variable, $\boldsymbol{F}_H(s) \in \mathbb{R}^{3}$ is the Laplace transform of the measured forces and $\boldsymbol{V}_{adm}(s) \in \mathbb{R}^{3}$ is the Laplace transform of the admittance reference translational velocity.

\subsubsection{Reference Generator}

As mentioned earlier, the main contribution of this work is to enable human-robot co-transportation of objects having different deformability. Due to such property of the object, relying just on haptic information may not be sufficient for an effective coordination of the co-carrying task. To address this issue, we introduce the \emph{Reference Generator}, which computes a velocity reference ($\boldsymbol{{v}}_{d}$) based on the admittance reference velocity ($\boldsymbol{{v}}_{adm}$) and the human hand velocity ($\boldsymbol{{v}}_{h}$) measured by the MoCap system. This unit calculates an adaptive index during the collaboration as follows:
\begin{align}
    \alpha =1-\frac{||\int_{t_{c}-W_{l}}^{t_{c}} \boldsymbol{{v}}_{adm}(t) \,dt||\ }{||\int_{t_{c}-W_{l}}^{t_{c}} \boldsymbol{{v}}_{h}(t) \,dt||\  + \epsilon} ,
\end{align}
\noindent where $\alpha \in [0,1]$ is the adaptive index, $t_c$ is the current time, $W_l$ is the length of the sliding time window and $\epsilon$ is a small number used to avoid the problem of division by zero. This index allows to understand whether the object is non-deformable ($\alpha = 0$), deformable ($\alpha = 1$) or partially deformable ($\alpha \in (0,1)$). Note that,  $\alpha$ is  saturated at 0.

Subsequently, $\alpha$ is used on-the-fly to regulate the contribution of $\boldsymbol{{v}}_{h}$ to $\boldsymbol{{v}}_{d}$ using:
\begin{align}
    \boldsymbol{{v}}_{d} = \boldsymbol{{v}}_{adm} + \alpha\boldsymbol{{v}}_{h}.
\end{align}

This formulation allows to benefit from two different sources of information. 
In order to clarify how the \textit{Reference Generator} deals with objects having different deformation properties, three distinct cases can be analyzed. If the object being carried does not transmit any force applied by the human (e.g. a loose rope), $\alpha$ is closer to 1.  In that case, $\boldsymbol{{v}}_{adm}$ is always close to zero and consequently, $\boldsymbol{{v}}_{d} \approx \boldsymbol{{v}}_{h}$. On the other hand, if the carried object is capable of transferring the forces applied by the human (e.g., a rigid box), human and robot will move rigidly together ($\boldsymbol{{v}}_{h} \approx \boldsymbol{{v}}_{adm}$). In this situation, $\alpha$ is closer to 0, resulting in $\boldsymbol{{v}}_{d} \approx \boldsymbol{{v}}_{adm}$. The two examples presented so far are the extreme cases of deformable and rigid objects, respectively. When the objects exhibit varying characteristics in terms of deformability, $\alpha$ will take values between 0 and 1, resulting in a reference  velocity computed by a combination of $\boldsymbol{{v}}_{adm}$ and $\boldsymbol{{v}}_{h}$.

Finally, the actual reference pose and twist are sent to the Whole-Body Controller and they are computed as follows:
\begin{align}
    \boldsymbol{{x}}_{d}= \int_{0}^{t_{c}} \boldsymbol{\dot{x}}_{d}(t) \,dt\ ,
\end{align}
\noindent where $\boldsymbol{\dot{x}}_{d} = [\boldsymbol{v}_{d}^{T}, \boldsymbol{0}^{T}]^T$.

\section{EXPERIMENTS}

To verify the effectiveness of our proposed framework, we compared the performance of our Adaptive Collaborative Interface (described in Section~\ref{sec:methodology}) to an admittance controller during a human-robot co-carrying task of a partially deformable object.

\subsection{Experimental Setup}

In this work, the robotic platform Kairos (see Fig.~\ref{fig:system_overview_fig}) was used for our experiments. It consists of an Omni-directional Robotnik SUMMIT-XL STEEL mobile base, and a high-payload (16 kg) 6-DoFs Universal Robot UR16e arm attached on top of the base. In addition, an underactuated Pisa/IIT SoftHand was mounted at the end-effector of the robotic arm to grasp the object being carried.

The Xsens (Fig.~\ref{fig:system_overview_fig}) was used as MoCap system to track human movements. It is composed of seventeen Inertial Measurement Units (IMUs) placed on specific parts of the human body. Thanks to this system, it is possible to get real-time measurement of the human hand, that is fed to our controller as described in Section~\ref{sec:methodology}. In order to analyze the movements of the object (only for validation purposes), we decided to utilize OptiTrack MoCap system. To this end, two different marker sets were placed near the human and robot grasping positions of the object. To emphasize, the data collected from this system was only used for the analysis of the experiments.

The muscular activity of Anterior Deltoid (AD), Posterior Deltoid (PD), Biceps Brachii (BB), Triceps Brachii (TB), Flexor Carpi (FC), Extensor Digitorum (ED), Erector Spinae (ES) longissimus, and Multifidus (MF) was recorded using the Delsys Trigno platform, a wireless sEMG system commercialized by Delsys Inc. (Natick, MA, United States). The sEMG sensors were placed on the selected muscle groups of each participant's right arm and right part of the back according to SENIAM recommendations \cite{HERMENS2000}. Afterward, the signals obtained were filtered and normalized with their Maximum Voluntary Contractions (MVC).

\subsection{Experimental Procedure}

\begin{figure}
    \centering
    \resizebox{0.9\columnwidth}{!}{\rotatebox{0}{\includegraphics[trim=3cm 0.5cm 4cm 3cm, clip=true]{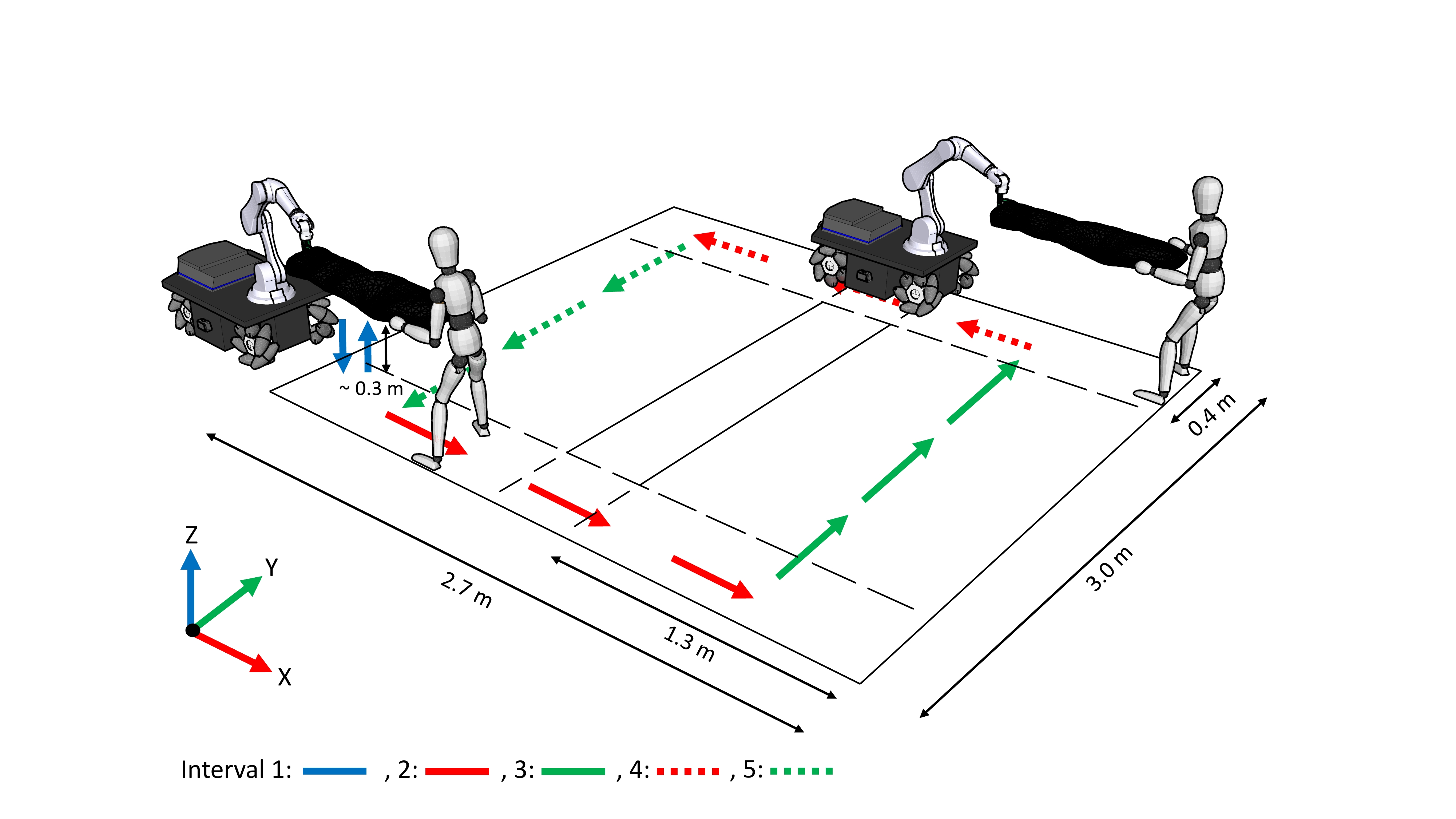}}}
    \caption{Illustration of the experiment and its intervals, in which an object is co-transported with a mobile manipulator along a path to be followed.}
    \vspace{-0.7cm}
    \label{fig:trajectory_and_object}
\end{figure}

In the experiment, the participants were asked to co-carry a partially deformable object (see Fig.~\ref{fig:system_overview_fig}) along a designed path in collaboration with the robot. The path comprises six main sub-movements: an initial down-up movement (2 of 6) followed by a square trajectory on a plane (4 of 6) parallel to the floor at a height that is customized based on the preference of each subject, as shown in Fig.~\ref{fig:trajectory_and_object}. The object features an approximately rigid behaviour if stretched along its larger dimension, while it is rather deformable if stressed along the other directions. The following guidelines were given to the participants before the beginning of the experiment:

\begin{itemize}
    \item The object must be kept inside the drawn path.
    \item If the object is tilted at the end of each sub-movement, the participant cannot start the following one. In this case, the participant must transmit a force to the robot by deforming the object to align the robot to the intended position and then start the following sub-movement.
    \item The task must be completed as fast as possible.
\end{itemize}

Each participant performed a total of 10 trials (2 controllers $\times$ 5 repetitions). The order of the trials was randomized not to affect the results with the learning effect. Before the experiment, a familiarization phase for both controllers was conducted until the subjects felt at their ease with the system. To guarantee human safety, mobile base velocities and the arm forces were constantly monitored and limited.

\subsection{Participants}

Twelve healthy volunteers, six males and six females, (age: $26.8 \pm 6.7$ years; mass: $64.9 \pm 16.2$ kg; height: $172.4 \pm 10.1$ cm)\footnote{Subject data is reported as: mean $\pm$ standard deviation.} were recruited for the experiments. After explaining the experimental procedure, written informed consent was obtained, and a numerical ID was assigned to anonymize the data. The whole experimental activity was carried out at the Human-Robot Interfaces and Physical Interaction (HRII) Lab, Istituto Italiano di Tecnologia (IIT), in accordance with the Declaration of Helsinki. The protocol was approved by the ethics committee Azienda Sanitaria Locale (ASL) Genovese N.3 (Protocol IIT\_HRII\_ERGOLEAN 156/2020).

\subsection{Controller Parameters}

In order to have a fair comparison between the ACI and the standard admittance controller, the same desired mass ($\boldsymbol{M}_{adm} = diag \{6,6,6\}$) and damping ($\boldsymbol{D}_{adm} = diag \{30,30,30\}$) were assigned for both controllers in this study. These parameters were selected based on a heuristic, in order to balance a trade-off between transparency and stability during the co-transportation.

Furthermore, the length of the sliding time window ($W_l$) which was used for the calculation of the adaptive index ($\alpha$) was selected as 0.25 s. This time length was tuned in order to compromise the delay in the detection of a change in the object deformability with its accurate identification. 

Regarding the Whole-Body Controller, $\boldsymbol{K}$, $\boldsymbol{W}_1$ and $\boldsymbol{W}_2$ were experimentally tuned so that an accurate tracking of the desired motion was achieved while avoiding too high joint velocities. The values selected for this purpose are $\boldsymbol{K}=diag\{1.0,1.0,1.0,0.1,0.1,0.1\}$, $\boldsymbol{W}_1=100\cdot diag\{10,10,10,5,5,5\}$ and $\boldsymbol{W}_2=3\cdot diag\{\boldsymbol{1}_m\}$. Moreover, in order to guarantee a locomotion behaviour (base following arm movements), $\boldsymbol{W}_3=diag\{\boldsymbol{0}_{n_b},\boldsymbol{1}_{n_a}\}$.

\subsection{Performance Metrics and Assessment Tools}
\label{performance_metrics_section}

\begin{figure*}[!t]
 \centering
    {\includegraphics[width=0.8\textwidth, trim=4cm 0cm 4cm 0cm, clip=true]{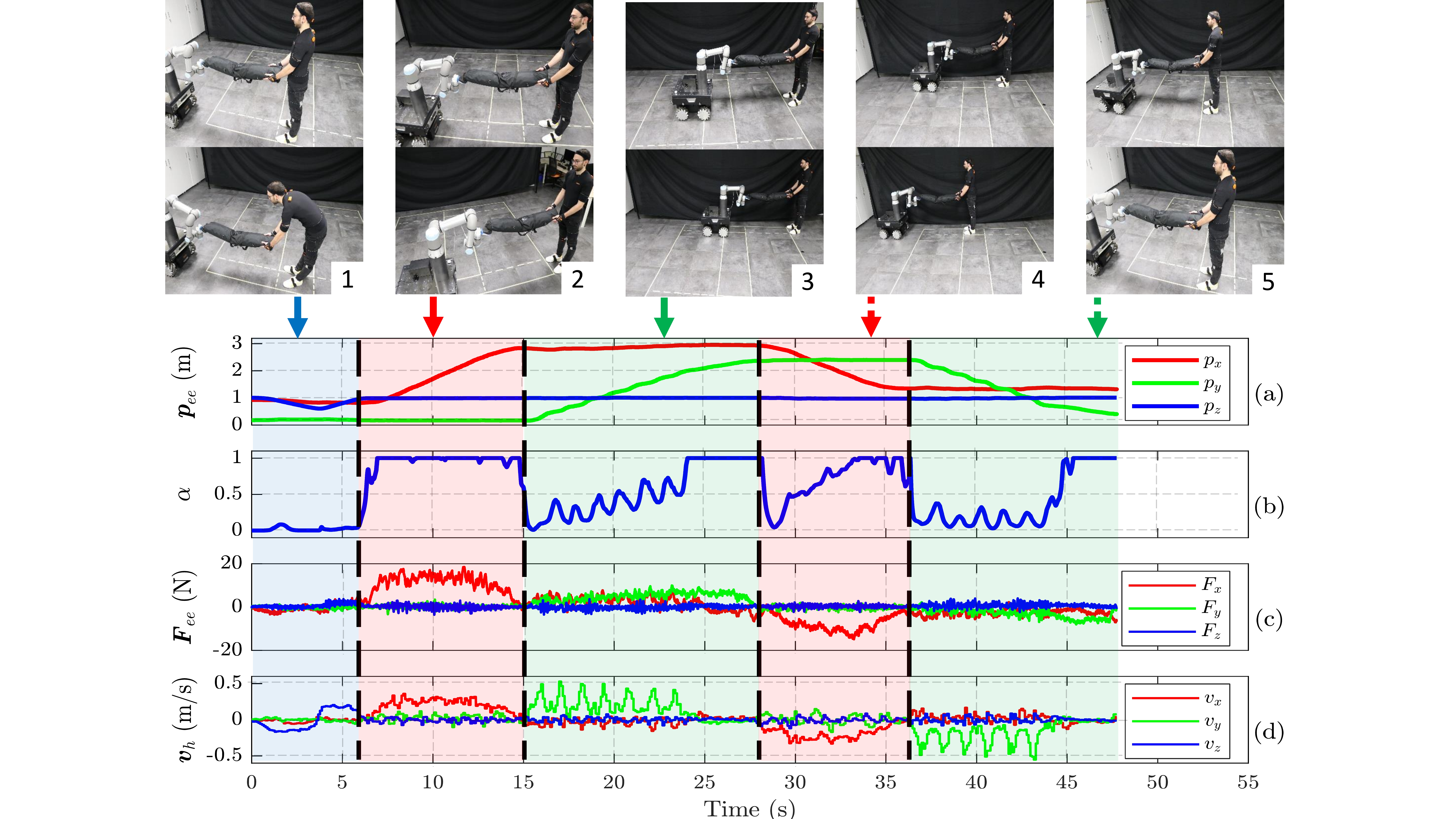}}
    \caption{Snapshots of the experiment: (1) Lowering and lifting; (2) Pulling; (3) Moving sideways to the right; (4) Pushing; and (5) Moving sideways to the left. Lower plots show the values obtained with the proposed controller during the collaborative transportation task of a typical trial.}
    \vspace{-0.7cm}
    \label{fig:intervals}
\end{figure*}

After the experiments, the participants filled a questionnaire to rate different qualitative aspects of their experience for each controller. The questionnaire is composed of a standard part, namely the NASA-TLX \cite{HART1988139},  and a custom Likert scale-based part designed specifically for this study. The custom part of the questionnaire includes 10 sentences. Q.1 I felt safe during the cooperation with the robot; Q.2 The robot understood my intention easily; Q.3 It was easy to keep the object inside the track; Q.4 The robot made the co-carry comfortable for me; Q.5 The robot's actions were harmless; Q.6 I was confident with the robot during the task; Q.7 I would choose the robot as an assistant to accomplish the task; Q.8 The robot was helping me during the task; Q.9 The robot was reliable during the task; Q.10 The performance of the controller in different directions of the co-carrying task was quite similar. Besides, for the quantitative evaluation of the task-related performance, the following metrics were used.

\begin{itemize}

    \item \textbf{Completion Time ($\boldsymbol{t_{c}}$)}: This is the elapsed time from the beginning to the end of the co-carrying task.
    
    \item \textbf{Alignment ($\boldsymbol{D_{AM}}$)}: This metric is formulated to calculate the robot performance in following human movements during the co-carrying task by taking into account the alignment of the object between human and robot. It is computed in the following way:

    \small
    \begin{equation}
    \begin{aligned}
        D_{AM}= \frac{\int_{t_{s}}^{t_{e}}{||\boldsymbol{r}_{crm}(t) - \boldsymbol{r}_{chm}(t) - (\boldsymbol{r}_{srm}- \boldsymbol{r}_{shm}) ||dt}}{t_{e}-t_{s}},
    \end{aligned}
    \end{equation}    
    \normalsize
    
    where $\boldsymbol{r}_{chm}$ and $\boldsymbol{r}_{crm}$ are the current human and robot marker positions, $\boldsymbol{r}_{shm}$ and $\boldsymbol{r}_{srm}$ are human and robot marker positions at the beginning of the experiment, and $t_s$ and $t_e$ indicate the starting and ending time of the experiment. 
    
    \item \textbf{Effort ($\boldsymbol{E_{EMG}}$)}:
    The effort is inferred through the measured muscle activities of the participants' selected muscle groups. It is obtained for each individual muscle by calculating the average of the processed sEMG data during the experiment.
\end{itemize}

\section{RESULTS AND DISCUSSION}

For a deep analysis of the experimental results during the different movement directions, the experiment has been divided into 5 intervals as depicted in Fig.~\ref{fig:intervals}: (1) Lowering and lifting; (2) Pulling; (3) Moving sideways to the right; (4) Pushing; and (5) Moving sideways to the left. In the same figure, the results obtained from an example experiment with the proposed framework are shown.

\begin{figure}[!t]
    \centering
    {\includegraphics[width=0.88\columnwidth, trim=7.1cm 9.8cm 16.5cm 1.6cm, clip=true]{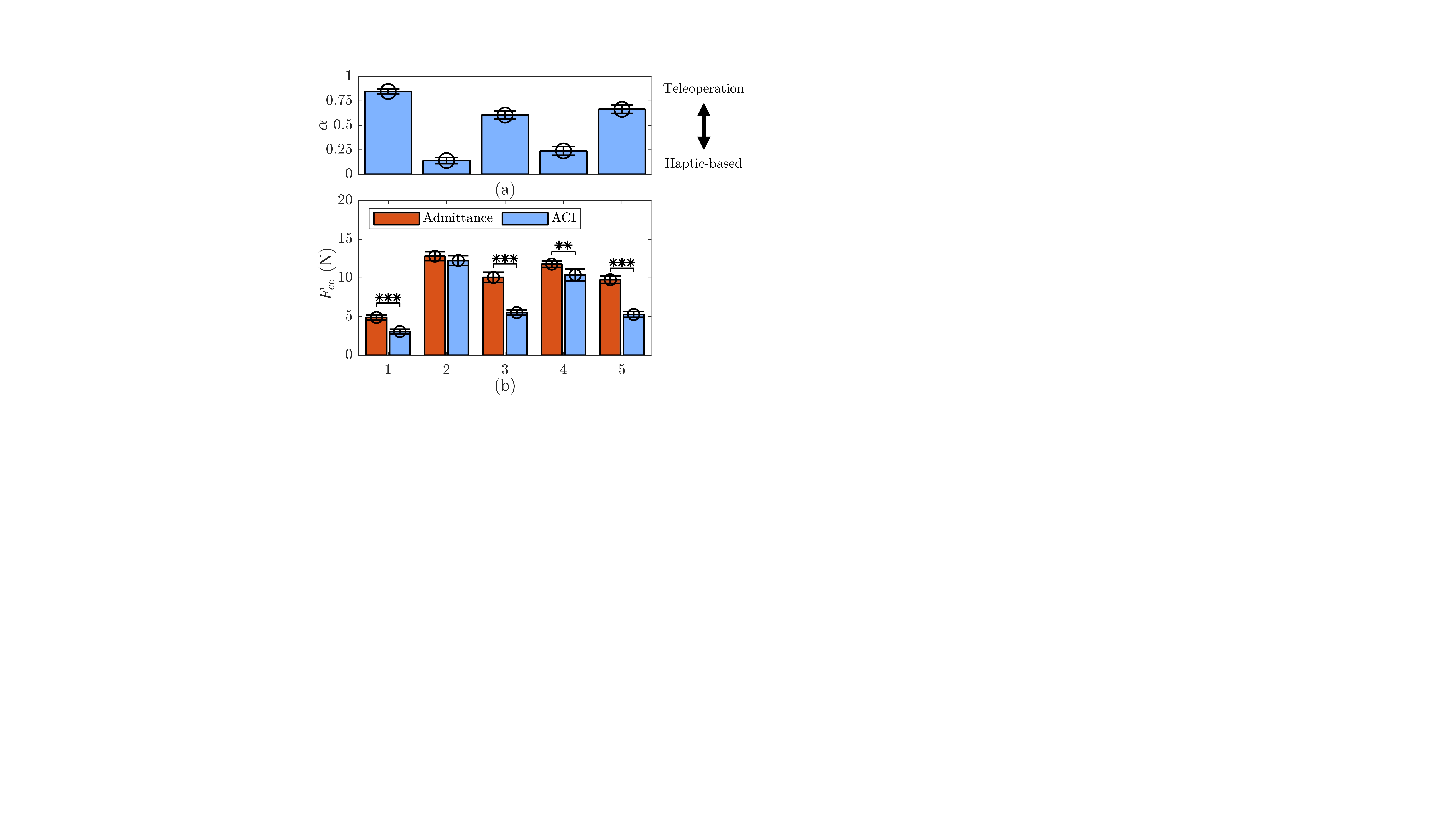}}
    \caption{(a) The means and standard errors of the adaptive index, $\alpha$, during the intervals, (b) the measured force amplitude from the end-effector of the robot for both controllers during the intervals along with outcomes of sign-test carried out: *: p $< 0.05$, **: p $< 0.01$, ***: p $< 0.001$.}
    \vspace{-0.7cm}

    \label{fig:interval_performance}
\end{figure}

As mentioned in Sec.~\ref{sec:methodology}, when the adaptive index ($\alpha$) approaches 1, it shows that the object being carried is highly deformable. Vice-versa, when $\alpha$ approaches 0, this indicates the rigidity of the object. Fig.~\ref{fig:interval_performance}a shows the means and standard errors of $\alpha$ during the intervals, which are calculated by averaging for all participants. As it can be seen, the highest mean is observed during the first interval, where it is close to 1. Conversely, the smallest one is observed during the second one among all intervals, where it approaches 0. Moreover, the third, fifth and fourth intervals exhibit intermediate values, with means of 0.6 for the former two and 0.25 for the latter. These results indicate that the object in combination with the grasping type used, is not rigid while lowering and lifting, and it behaves as a non-deformable object when it is fully stretched while pulling. Additionally, the object is partially deformable when it is pushed and moved sideways. Note that, similar deformability is estimated in the two directions of sideways movements. These outcomes are expected considering the characteristics of the object being carried combined with the non-purely rigid Pisa/IIT Softhand placed at the end-effector.

\begin{figure}[!b]
    \centering
    {\includegraphics[width=0.75\columnwidth, trim=0cm 0cm 0cm 0cm, clip=true]{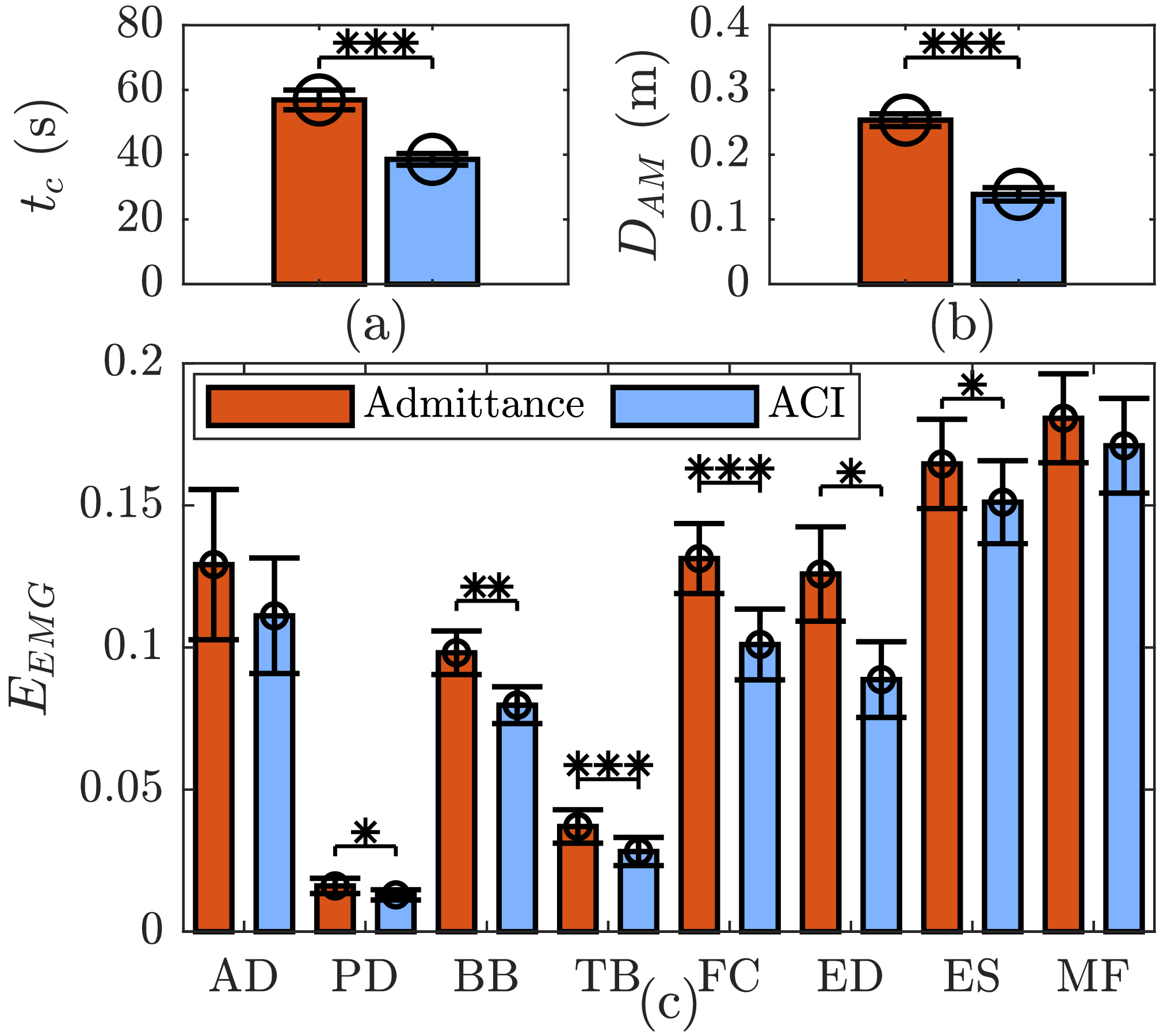}}
    \caption{The means and the standard errors for the performance metrics along with the outcomes of the sign-test carried out for the compared controllers: *: p $< 0.05$, **: p $< 0.01$, ***: p $< 0.001$.}
    \vspace{-0.7cm}
    \label{fig:performance_metrics}
\end{figure}  

The means and standard errors of the force amplitude measured at the robot's end-effector for the two controllers during the intervals are depicted in Fig.~\ref{fig:interval_performance}b. These results are calculated by finding the average of them for all participants. When these results are analyzed, it can be seen that the admittance controller presents statistically significant higher force amplitudes according to sign-test for all intervals except the second one. This difference is due to the deformability of the object, which is indicated by high values of $\alpha$. These high values can be interpreted as the difficulty of transferring forces from the human to the robot through the object. The ACI copes with this issue by estimating humans intention by means of the velocity of their hand. Instead, the admittance controller fails, and the cooperating human needs to deform the object to allow the force transmission and the intention communication. Notice that, in the second interval no significant differences are revealed between the two controllers, as expected. Indeed, in that segment of the task the object is almost purely rigid (i.e. $\alpha$ approaches 1).

Fig.~\ref{fig:performance_metrics} illustrates the means and standard errors of the performance metrics described in subsection~\ref{performance_metrics_section} for both the proposed controller and the admittance controller. The performance metrics are obtained by averaging these measured quantities for all participants. Sign-tests are performed to evaluate the effect of the controller on the performance metrics and the outcomes of them can be found (see Fig.~\ref{fig:performance_metrics}). The results show that the completion time and the alignment for the proposed controller are significantly lower (p $< 0.001$) than those of the admittance controller (see Fig.~\ref{fig:performance_metrics}a, \ref{fig:performance_metrics}b). The mean efforts for all the selected muscle groups are lower under the proposed controller (see Fig.~\ref{fig:performance_metrics}c). However, there is no significant effect of controller type on the effort of AD and MF muscles. These results indicate that participants completed the task faster, with less effort and higher effectiveness under the proposed controller.

\begin{figure}[!t]
    \centering
    {\includegraphics[width=0.75\columnwidth, trim=0cm 0cm 0cm 0cm, clip=true]{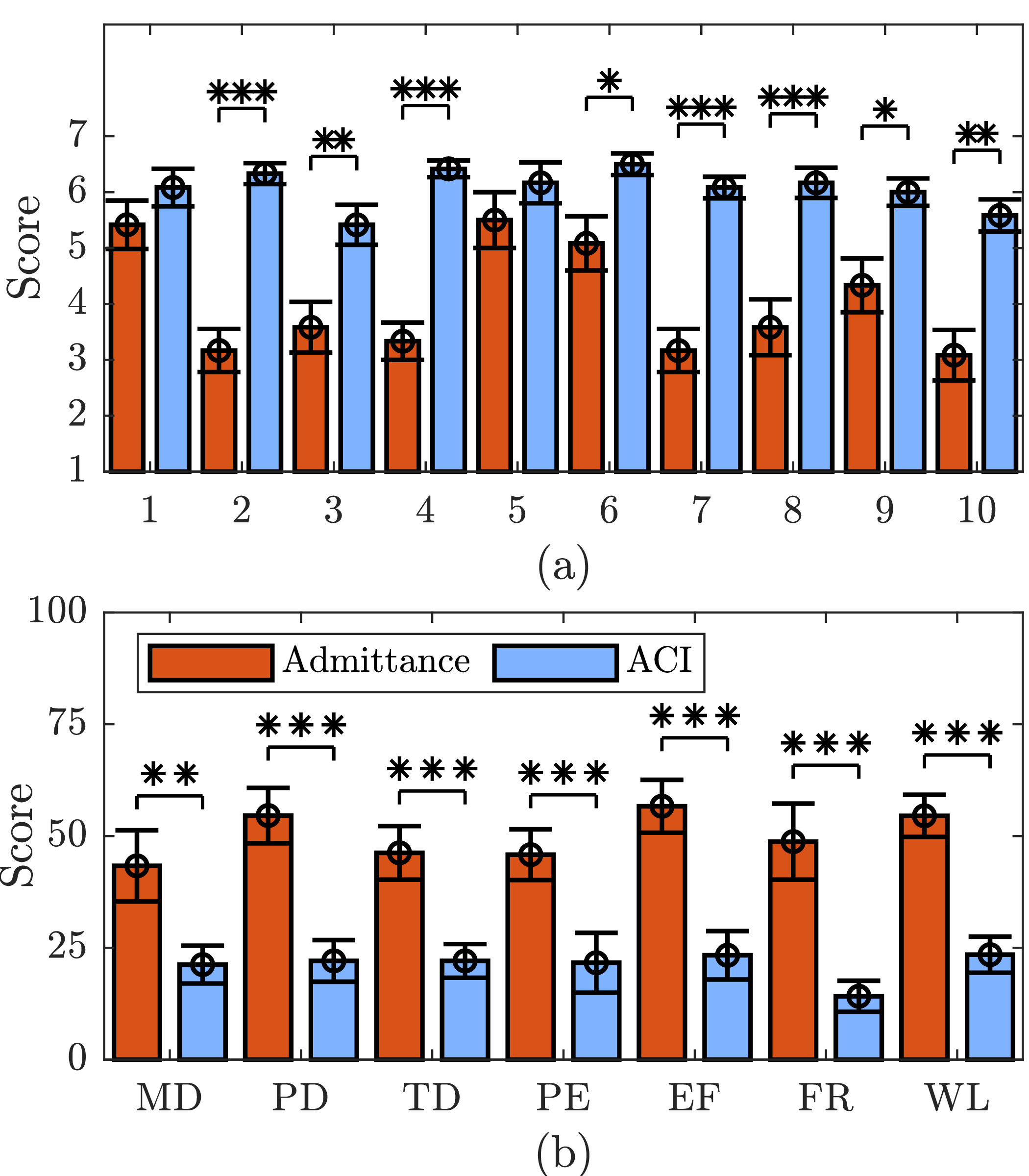}}
    \caption{The means and the standard errors obtained from the custom (a) and NASA-TLX (b) questionnaires for the two controllers along with the outcomes of the sign-test carried out for the compared controllers: *: p $< 0.05$, **: p $< 0.01$, ***: p $< 0.001$.}
    \vspace{-0.7cm}
    \label{fig:questionnaires}
\end{figure}

The qualitative results are reported in Fig.~\ref{fig:questionnaires}. For each questionnaire statement, the means and standard errors for both controllers are represented as bar plots, and the outcomes of the sign-tests are reported. Particularly, Fig.~\ref{fig:questionnaires}a depicts the outcomes of the custom questionnaire, which aims to obtain a subjective evaluation of the two controllers, while Fig.~\ref{fig:questionnaires}b shows the results obtained for the NASA-TLX questionnaire through which the participants score their perceived workload for six subscales. The statements of the custom questionnaire can be classified into 3 categories: safety (Q.1 and 5); performance (Q.2, 3, 8, and 10); and usability (Q.4, 6, 7, and 9). As shown in Fig.~\ref{fig:questionnaires}a, the proposed controller has better score means compared with the admittance controller in all statements of the custom questionnaire. In addition, statistically significant differences (p $< 0.05$) are observed for all statements apart from the ones related to safety. These results demonstrate that the proposed controller surpasses the admittance controller in terms of performance and usability according to the participants while ensuring safety which is of utmost importance for a controller in pHRI. Regarding the NASA-TLX questionnaire (see Fig.~\ref{fig:questionnaires}b), the ACI has statistically significant superior scores in all the categories, as well as in the computed overall workload.

\section{FRAMEWORK APPLICABILITY}

In the previous sections, the applicability of the ACI has been demonstrated with the object used in our experimental setup. In order to validate the proposed controller can also handle objects with different deformability, we tested two extreme cases (a rope and an aluminum profile) as illustrated in Fig.~\ref{fig:digest_figure}. Thanks to the adaptive features of ACI, both objects could be co-transported successfully. These experiments are demonstrated in the multimedia attachment\footnote{The video can also be found at \url{https://youtu.be/ounQmCdYlC4}}.

\section{CONCLUSION}
This paper presented an adaptive control framework that can handle objects with unknown deformability for human-robot collaborative transportation. Instead of relying just on the deficient haptic information that is transferred through the deformable object, the proposed controller allows us to benefit from the information of human movements. Based on this perceived sensory information, it is also possible to react to changes in deformability of the object through an adaptive index computed online. 

The effectiveness of the presented framework was evaluated experimentally by comparing it to an admittance controller during a collaborative carrying task. A thorough analysis based on quantitative and qualitative results was carried out to compare the performance of the two controllers. The results showed that the proposed controller is more successful than the admittance controller on providing assistance during co-carrying of deformable objects.  

In the future, we plan to tackle the remaining challenge of co-transportation, which is rotation/translation ambiguity, by employing the kinematic information of the human body. Moreover, we will focus on replacing our human motion capture system (Xsens suit) with a less expensive vision-based system. With this change, we will make our framework more suitable for the industry by overcoming disadvantages such as the need of wearing a suit.

\bibliographystyle{ieeetr}
\bibliography{biblio.bib}

\end{document}